\documentclass[letterpaper, 10 pt, conference]{ieeeconf} 

\IEEEoverridecommandlockouts                              
\usepackage{graphicx,subfigure}
\usepackage{epsfig} 
\usepackage{mathptmx} 
\usepackage{times} 
\usepackage{amsmath} 
\usepackage{amssymb}  
\usepackage{dsfont}
\usepackage{verbatim}

\usepackage{pgfplots}\pgfplotsset{compat=1.4}

\usepackage{graphicx}
\usepackage{hyperref}

\numberwithin{equation}{section}

\newtheorem{lemma}{Lemma}[section]
\newtheorem{remark}[lemma]{Remark}
\newtheorem{theorem}[lemma]{Theorem}
\newtheorem{corollary}[lemma]{Corollary}

\newtheorem{definition}[lemma]{Definition}
\newcommand{\mB}[1]{{\mathbb{#1}}}

\newcommand{\half}{\mbox{\small$\frac{1}{2}$}}

\newcommand{\SKp}{\mathcal{S}^K_+}

\newcommand{\B}[1]{{\bf #1}}

\newcommand{\R}{{\mathbb R}}

\newcommand{\F}{\mathcal{F}}

\begin{document}

\title{\LARGE \bf Linear system identification\\ using stable spline kernels and PLQ penalties}

\author{Aleksandr Y.~Aravkin, 
        James V.~Burke 
       and~Gianluigi~Pillonetto
        \thanks{A.Y. Aravkin (saravkin@us.ibm.com) is with IBM T.J. Watson Research Center, Yorktown Heights, NY, 10598}
                \thanks{J.V. Burke (burke@math.washington.edu) is with Department of Mathematics, University of Washington, Seattle, USA}
        \thanks{G. Pillonetto (giapi@dei.unipd.it) is with Dipartimento di Ingegneria
        dell'Informazione, University of Padova, Padova, Italy.}
        \thanks{This research has been partially supported 
by the European Community under
agreement n. FP7-ICT-223866-FeedNetBack, n257462 HYCON2 Network of excellence,
 by the FIRB project entitled ``Learning meets time",
and by the advanced grant LEARN from the 
European Research Council, contract 267381.}}

\maketitle \thispagestyle{empty} \pagestyle{empty}

\begin{abstract}
%
The classical approach to linear system identification
is given by parametric Prediction Error Methods (PEM).
In this context, model complexity is often unknown
so that a model order selection step is needed
to suitably trade-off bias and variance.
Recently, a different approach to linear system identification has been
introduced, where model order determination is avoided 
by using a regularized least squares framework.
In particular, the penalty term on the impulse response
is defined by so called {\it stable spline kernels}.
They embed information on regularity and BIBO stability, and
depend on a small number of parameters which can be estimated from data.
In this paper, we provide new nonsmooth formulations of the stable spline estimator.
In particular, we consider linear system identification problems in a very broad context, 
where regularization functionals and data misfits can come from a rich set of 
{\bf piecewise linear quadratic} functions. Moreover, our analysis includes 
polyhedral inequality constraints on the unknown impulse response. 
For any formulation in this class, we show that interior point methods can be 
used to solve the system identification problem, with complexity $O(n^3) + O(mn^2)$
in each iteration, where $n$ and $m$ are the number of 
impulse response coefficients and measurements, respectively. 
The usefulness of the framework is illustrated
via a numerical experiment where output measurements
are contaminated by outliers.
\end{abstract}
\begin{keywords}
linear system identification; bias-variance trade off; kernel-based
regularization; robust statistics; interior point methods;
piecewise linear quadratic densities 
\end{keywords}

\section{Introduction}

The classical approach to linear system identification
is given by Parametric Prediction Error Methods (PEM)
\cite{Ljung,Soderstrom}. First, models 
of different and unknown order, e.g. ARX or ARMAX, are postulated
and identified from data. Then, they are compared using either complexity 
measures such as AIC or cross validation (CV) \cite{Akaike1974,Hastie01}.\\
Some limitations of this approach have been recently described in
\cite{SS2010} (see also \cite{PitfallsCV12}
for an analysis of CV). 
This has led to the introduction of an alternative technique,
where identification is seen as a function learning problem formulated
in a possibly infinite-dimensional space
 \cite{SS2010,SS2011}. In particular, the problem
 is cast in the framework of Gaussian regression 
\cite{Rasmussen}: the unknown impulse response
is seen as a Gaussian process, whose autocovariance
 encodes available prior knowledge. This approach was subsequently
 given an interpretation in a Regularized Least Squares framework in  \cite{ChenOL12}.\\ 
The new estimators proposed in \cite{SS2010,PillACC2010}
rely on a class of autocovariances, called stable spline kernels, 
which include information on the exponential stability of the unknown system.
The impulse response is modeled as the $m$-fold
integration of white Gaussian noise subject to an exponential time transformation.
The first-order stable spline kernel has been recently derived
using {\it deterministic} arguments  \cite{ChenOL12}, and named the TC kernel.  
An even more sophisticated 
covariance for system identification, the so called DC kernel, is also described in
\cite{ChenOL12}.\\
All of these kernels are defined by a small number of unknown hyperparameters, 
which can be learned from data,
e.g. by optimizing the marginal likelihood \cite{Maritz:1989,MacKay,BergerBook}.
This procedure resembles model order selection in the classical parametric paradigm,
and theoretical arguments supporting it are illustrated in \cite{AravkinIFAC12}.
Once the hyperparameters are found, the estimate of the system impulse response becomes
available in closed form. Extensive simulation studies have shown that 
these new estimators can lead to significant advantages with respect
to the classical ones, in particular in terms of robustness and 
in model complexity selection.\\
All of the new kernel-based approaches discussed in
 \cite{SS2010,SS2011,ChenOL12} rely on quadratic loss and
and penalty functions. As a result,
in some circumstances they may
perform poorly. In fact, quadratic penalties 
are not robust when outliers are present in the data
\cite{Hub,Gao2008,Aravkin2011tac,Farahmand2011}. 
In addition, they neither promote sparse solutions,
nor select small subsets of measurements or 
impulse response coefficients  with the 
greatest impact on the predictive capability for
future data. These are key issues for
feature selection and compressed sensing
\cite{Hastie90,LARS2004,Donoho2006}.\\
The limitations of quadratic penalties motivate
adopting alternative penalties for both loss and regularization functionals. 
For example,
popular regularizers are the
the $\ell_1$-norm, as in the LASSO
\cite{Lasso1996}, or a weighted combination of $\ell_1$ and 
$\ell_2$, as in
the elastic net procedure~\cite{EN_2005}.
Popular fitting measures robust to outliers are the $\ell_1$-norm, the Huber loss~\cite{Hub}, 
the Vapnik $\epsilon$-insensitive loss~\cite{Vapnik98,Pontil98} 
and the hinge loss~\cite{Evgeniou99,Pontil98,Scholkopf00}.
Recently,  all of these approaches have been cast in a unified 
statistical modeling framework~\cite{AravkinIFAC12,AravkinCDC12}, 
where solutions to all models 
can be computed using interior point (IP) methods.\\
The aim of this paper is to extend this framework 
to the linear system identification problem. In particular, 
we propose new impulse response estimators
that combine the stable spline kernels and 
arbitrary piecewise linear quadratic (PLQ) penalties. 
Generalizing the work in
 \cite{AravkinIFAC12,AravkinCDC12}, we also allow the
 inclusion of inequality constraints  on the unknown parameters.
This generalization can be used to efficiently include additional information --- for example, about nonnegativity and unimodality of
the impulse response --- into the final estimate. 
We show that all of these
models can be solved with IP techniques, with complexity
that scales well with the number of output measurements.
These new identification procedures are tested via a Monte Carlo study
where output error models are randomly generated and output data 
(corrupted by outliers) is obtained. We compare the performance
of the classical stable spline estimator that uses a quadratic loss 
with the performance of the new estimator
that uses $\ell_1$ loss.\\
The structure of the paper is as follows. In Section \ref{Sec2}, 
we formulate the problem and briefly review the stable spline estimator 
described in \cite{SS2010,ChenOL12}. 
In Section \ref{Sec3} we introduce the new class of non smooth stable spline estimators,
review the class of PLQ penalties, and generalize the framework in~\cite{AravkinBurkePillonetto2013} 
by including affine inequality constraints. 
We also demonstrate how IP methods can be used to efficiently
compute the impulse response estimates.
In Section \ref{Sec4}, the new approach is tested 
via a Monte Carlo study, where system output measurements
are corrupted by outliers. We end the paper with Conclusions, and include
additional proofs in the Appendix.

\section{Problem statement and the stable spline estimator}\label{Sec2}

\subsection{Statement of the problem}


Consider the following linear time-invariant discrete-time system 
\begin{equation} \label{MeasMod}
y(t) = G(q)u(t)+e(t), \quad t=1,\ldots,m \;,
\end{equation}
where $y$ is the output, $q$ is the shift operator ($qu(t)=u(t+1)$), 
$G(q)$ is the linear operator associated with the true system, assumed stable, 
$u$ the input and $e$ 
the i.i.d. noise. Assuming the input $u$ known, our problem is to estimate the system
impulse response from $N$ noisy measurements
of $y$.

\subsection{The stable spline estimator}\label{SS+L2}

We now briefly review the regularized approach to system identification
proposed in \cite{SS2010,ChenOL12}. 
For this purpose, denote by $x \in \R^n$ the (column) vector 
containing the impulse response coefficients. Here, in contrast to classical
approaches to system identification, the size $n$ is 
chosen sufficiently large to capture system 
dynamics rather than 
to establish any kind of trade-off between 
bias and variance. 
It is useful to rewrite the measurement model
(\ref{MeasMod}) using the following matrix-vector notation
\begin{equation}\label{MatrixMod}
  z=H x + E\;,  
\end{equation}
where the vector $z \in \R^m$ contains the $m$ output measurements, 
$H$ is a suitable matrix defined by input values, and $E$ denotes the noise
of unknown variance $\sigma^2$. 
Then,  the stable spline estimator is defined by the following
regularized least squares problem:
\begin{equation}
\label{eq:MV}
  \hat x = \arg \min_{x} \|z-H x\|_2^2  + \gamma x^T Q^{-1} x\;,
\end{equation}
where the positive scalar $\gamma$ is a regularization parameter,
and $Q \in \R^{n \times n}$ can be taken from the class of stable
spline kernels \cite{PillACC2010}. 
In particular, adopting the discrete-time version 
of the stable spline kernel of order 1,   
the  $(i,j)$ entry of $Q$ is
\begin{equation}\label{eq:TC}
Q_{ij} = \alpha^{\max(i,j)}, \quad 0 \leq \alpha <1\;.
\end{equation} 
Above, $\alpha$ is a kernel hyperparameter which corresponds
to the dominant pole of the system, and is typically unknown.
This kernel was also studied in
\cite{ChenOL12}, where it was called the tuned/correlated (TC) kernel.
Motivations underlying the particular shape (\ref{eq:TC}) have been 
discussed under both a statistical and a deterministic framework,
see  \cite{CDC2011P1} and \cite{CDC2011P2}.\\
Note that  the estimator (\ref{eq:MV}), equipped
with the kernel (\ref{eq:TC}), contains the 
unknown hyperparameters $\alpha$ and $\gamma$.
These can be obtained as follows.
First, the estimate $\hat\sigma^2$ of $\sigma^2$  can be computed by fitting  
a low-bias model for the impulse response using least squares 
(as e.g. described in \cite{Goodwin1992}).
Then, one can exploit the Bayesian interpretation
underlying problem (\ref{eq:MV}): if the noise is Gaussian, it provides the minimum
variance estimate of $x$ when the impulse response is modeled as 
a Gaussian vector independent of $E$ with autocovariance $\lambda Q$.
Here, $\lambda$ is an unknown scale factor equal to $\sigma^2 / \gamma $.
The estimates of $\lambda$ and $\alpha$ are obtained
by maximizing the marginal likelihood (obtained by integrating $x$ out 
of the joint density of $z$ and $x$). This gives 
\begin{equation}\label{eq:marglik}
(\hat \lambda, \hat \alpha) 
=
\arg \min_{\lambda,\alpha} \; z^T \Sigma^{-1}_z z + \log \det (\Sigma_z)\;,
\end{equation}
where the $m \times m$ matrix $\Sigma_z$ is 
$$
\Sigma_z = \lambda H Q H^T + \hat{\sigma}^2 I_m \;, 
$$ 
and $I_m$ the $m \times m$ identity matrix (see \cite{SS2010} for details).

Let $\hat Q$ be the matrix defined in (\ref{eq:TC}) 
with $\alpha$ set to its estimate $\hat \alpha$.
Then, setting $Q$ to $\hat Q$ and $\gamma$ to 
$\hat{\sigma}^2 / \hat{\lambda}$ in (\ref{eq:MV}),
we obtain the impulse response estimate 
$$
\hat x = \hat \lambda \hat Q H^T \hat \Sigma^{-1}_z z   \;,
$$ 
where
$$
\hat \Sigma_z = \hat \lambda H \hat Q H^T + \hat{\sigma}^2 I_m.  
$$

\section{New non smooth formulations of the stable spline estimator}\label{Sec3}


To simplify the problem formulation, it is useful 
to introduce an auxiliary variable $y$, and to  
to rewrite the classical stable spline estimator~\eqref{eq:MV}
using the following relationships:
\begin{equation}
\label{SysIdEqs}
x = Ly, \quad Q = LL^T.
\end{equation}
where $L$ is invertible. 
Using~\eqref{SysIdEqs},~\eqref{eq:MV}) becomes
\begin{equation}
\label{SysIdObjectiveTwo} 
\min_y
\left\|(z - H Ly)\right\|^2+ \gamma \|y\|^2\;.
\end{equation}
It is apparent that this estimator uses quadratic functions
to define both the loss $\left\|(z - H Ly)\right\|^2$
and the regularizer $\|y\|^2$.
In the rest of the paper we study a generalization
of (\ref{SysIdObjectiveTwo}) given by  
\begin{equation} 
\label{probTwo}
\min_{y \in Y} \quad  V \left(HLy -z\right)  + \gamma W\left(y\right) \;,
\end{equation}
where $Y$ is a polyhedral set 
(which can be used e.g. to provide nonnegativity information on the 
impulse response $x=Ly$), and $V$, $W$
are defined by the piecewise linear quadratic functions
introduced in the next subsection. 

\subsection{PLQ penalties}

\begin{figure}
\centering
\begin{tikzpicture}
  \begin{axis}[
    thick,
    width=.45\textwidth, height=2cm,
    xmin=-2,xmax=2,ymin=0,ymax=1,
    no markers,
    samples=50,
    axis lines*=left, 
    axis lines*=middle, 
    scale only axis,
    xtick={-1,1},
    xticklabels={},
    ytick={0},
    ] 
\addplot[domain=-2:+2]{.5*x^2};
  \end{axis}
\end{tikzpicture}
\begin{tikzpicture}
  \begin{axis}[
    thick,
    width=.45\textwidth, height=2cm,
    xmin=-2,xmax=2,ymin=0,ymax=1,
    no markers,
    samples=100,
    axis lines*=left, 
    axis lines*=middle, 
    scale only axis,
    xtick={-1,1},
    xticklabels={},
    ytick={0},
    ] 
  \addplot[domain=-2:+2]{abs(x)};
  \end{axis}
\end{tikzpicture}
\begin{tikzpicture}
  \begin{axis}[
    thick,
    width=.45\textwidth, height=2cm,
    xmin=-2,xmax=2,ymin=0,ymax=1,
    no markers,
    samples=50,
    axis lines*=left, 
    axis lines*=middle, 
    scale only axis,
    xtick={-1,1},
    xticklabels={$-\kappa$,$+\kappa$},
    ytick={0},
    ] 
\addplot[red,domain=-2:-1,densely dashed]{-x-.5};
\addplot[domain=-1:+1]{.5*x^2};
\addplot[red,domain=+1:+2,densely dashed]{x-.5};
\addplot[blue,mark=*,only marks] coordinates {(-1,.5) (1,.5)};
  \end{axis}
\end{tikzpicture}
\begin{tikzpicture}
  \begin{axis}[
    thick,
    width=.45\textwidth, height=2cm,
    xmin=-2,xmax=2,ymin=0,ymax=1,
    no markers,
    samples=50,
    axis lines*=left, 
    axis lines*=middle, 
    scale only axis,
    xtick={-0.5,0.5},
    xticklabels={$-\epsilon$,$+\epsilon$},
    ytick={0},
    ] 
    \addplot[red,domain=-2:-0.5,densely dashed] {-x-0.5};
    \addplot[domain=-0.5:+0.5] {0};
    \addplot[red,domain=+0.5:+2,densely dashed] {x-0.5};
    \addplot[blue,mark=*,only marks] coordinates {(-0.5,0) (0.5,0)};
  \end{axis}
\end{tikzpicture}
\begin{tikzpicture}
  \begin{axis}[
    thick,
    width=.45\textwidth, height=2cm,
    xmin=-2,xmax=2,ymin=0,ymax=1,
    no markers,
    samples=100,
    axis lines*=left, 
    axis lines*=middle, 
    scale only axis,
    xtick={-1,1},
    xticklabels={-1, 1},
    ytick={0},
    ] 
\addplot[domain=-2:+2]{.5*x^2 + 0.5*abs(x)};
  \end{axis}
\end{tikzpicture}
\begin{tikzpicture}
  \begin{axis}[
    thick,
    width=.45\textwidth, height=2cm,
    xmin=-2,xmax=2,ymin=0,ymax=1,
    no markers,
    samples=50,
    axis lines*=left, 
    axis lines*=middle, 
    scale only axis,
    xtick={-0.5,0.5},
    xticklabels={$-\epsilon$,$+\epsilon$},
    ytick={0},
    ] 
    \addplot[red,domain=-2:-0.5,densely dashed] {0.5*(-x-0.5)^2};
    \addplot[domain=-0.5:+0.5] {0};
    \addplot[red,domain=+0.5:+2,densely dashed] {0.5*(x-0.5)^2};
    \addplot[blue,mark=*,only marks] coordinates {(-0.5,0) (0.5,0)};
  \end{axis}
\end{tikzpicture}
 \caption{ Scalar penalties, top to bottom: $\ell_2$, $\ell_1$, Huber, 
 Vapnik, elastic net, and smooth insensitive loss}
\label{HuberVapnikFig}
\end{figure}
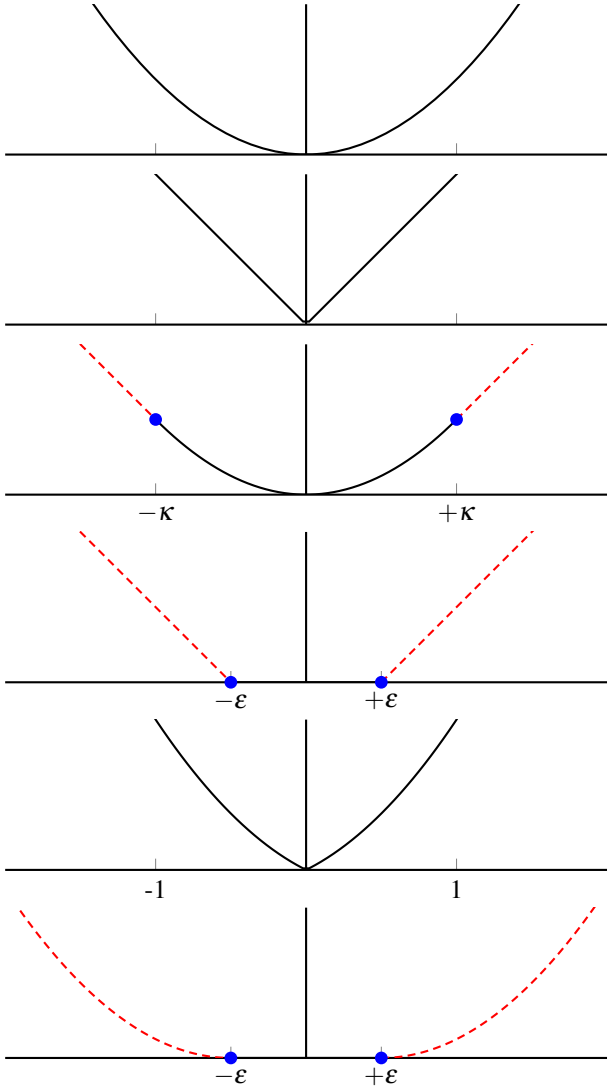

%
\begin{definition}[PLQ functions and penalties]
\label{generalPLQ}
A piecewise linear quadratic (PLQ) function is any function 
$\rho(U, M, b, B; \cdot): \mB{R}^N \rightarrow \mathbb{\overline R}$ 
having representation
\begin{equation}\label{PLQpenalty}
\rho(U, M, b, B; y) 
=
\sup_{u \in U}
\left\{ \langle u,b + By \rangle - \half\langle u, Mu
\rangle \right\} \;,
\end{equation}
where $U \subset \mB{R}^K$ is a nonempty polyhedral set, 
$M\in \SKp$ the set of real symmetric positive semidefinite matrices,
and $b + By$ is an injective affine transformation in $y$, with $B\in\mB{R}^{K\times N}$, 
so, in particular, $K \geq N$ and $\mathrm{null}(B) = \{0\}$. 

When $0 \in U$, the associated function is a {\it penalty}, since it 
is necessarily non-negative. 
\end{definition}
\begin{remark}
When $b = 0$ and $B = I$, we recover the basic piecewise linear-quadratic penalties 
characterized in \cite[Example 11.18]{RTRW}. 
\end{remark}

\begin{remark}[scalar examples]
\label{scalarExamples} 
$\ell_2$, $\ell_1$, elastic net, Huber, hinge, and Vapnik penalties 
are all representable using the notation of Definition
\ref{generalPLQ}.
\begin{enumerate}
\item\label{l2} $\ell_2$: Take $U = \mB{R}$, $M = 1$, $b = 0$, and $B = 1$. We obtain
\[ 
\displaystyle \rho(y) 
= \sup_{u \in \mB{R}}\left\{ uy -u^2/2 \right\}\;. 
\] 
The function inside the $\sup$ is
maximized at $u = y$, hence $\rho(y) = \frac{1}{2}y^2$. 

\item\label{l1} $\ell_1$: Take $U = [-1, 1]$, $M = 0$, $b = 0$, and $B = 1$. We obtain
\[ 
\displaystyle \rho(y) = \sup_{u \in [-1, 1]}\left\{ uy\right\}\;. 
\] The function inside the $\sup$ is maximized by
taking $u = \R{sign}(y)$, hence $\rho(y) = |y|$.

\item\label{elasticnet} Elastic net: $\ell_2 + \lambda \ell_1$. 
Take 
\[
U = \mB{R} \times [-\lambda, \lambda],\; 
b = \begin{bmatrix} 0 \\ 0 \end{bmatrix},\;
M = \begin{bmatrix} 1 & 0 \\ 0 & 0 \end{bmatrix},\;
B = \begin{bmatrix} 1 \\ 1\end{bmatrix}\;.
\]

\item\label{huber} Huber: Take $U = [-\kappa, \kappa]$, $M = 1$, $b = 0$, and $B = 1$.
We obtain 
\[ \displaystyle \rho(y) = \sup_{u \in [-\kappa, \kappa]}\left\{ uy - u^2/2 \right\}\;,\] 
with three explicit cases:
\begin{enumerate}
\item If $ y < -\kappa $, take $u = -\kappa$ to obtain
$-\kappa y  -\frac{1}{2}\kappa^2$.
\item If $-\kappa  \leq y \leq \kappa $, take $u = y$ to obtain
$\frac{1}{2}y^2$.
\item If $y > \kappa  $, take $u = \kappa $ to obtain
a contribution of $\kappa y -\frac{1}{2}\kappa^2$.
\end{enumerate}
This is the Huber penalty.


\item\label{vapnik} Vapnik loss is given by $(y-\epsilon)_+  + ( -y- \epsilon)_+$. 
We obtain its PLQ representation by  
taking 
\[
B = \begin{bmatrix}1 \\ -1\end{bmatrix},\;
b =  - \begin{bmatrix} \epsilon \\ \epsilon \end{bmatrix},\; 
M= \begin{bmatrix} 0 & 0 \\ 0 & 0\end{bmatrix}, \;
U = [0, 1]\times [0, 1]\;
\]
 to yield 
\[
\rho(y) 
= 
\sup_{u \in U}
\left\{
\left\langle \begin{bmatrix}y - \epsilon \\ 
-y - \epsilon \end{bmatrix} , u \right\rangle
\right\} 
= 
(y - \epsilon)_+ + (-y-\epsilon)_+. 
\]

%

\item \label{silf} Soft insensitive loss function~\cite{Chu01aunified}. 
We can create a symmetric soft insensitive loss function (which one might term the Hubnik) by 
adding together two soft hinge loss functions:  
\[
\begin{aligned}
\rho(y) &= \sup_{u \in [0,\kappa]}\left\{ (y - \epsilon) u\right\} -\half u^2 + \sup_{u \in [0,\kappa]}\left\{ (-y - \epsilon) u\right\} -\half u^2\\
& = \sup_{u \in [0,\kappa]^2}
\left\{
\left\langle \begin{bmatrix}y - \epsilon \\ 
-y - \epsilon \end{bmatrix} , u \right\rangle
\right\} 
- \half u^T \begin{bmatrix}1 & 0 \\ 0 & 1 \end{bmatrix}u\;.
\end{aligned}
\]
See bottom bottom panel of Fig.~\ref{HuberVapnikFig}.
\end{enumerate}
\end{remark}
%
%

\subsection{Optimization with PLQ penalties}
\label{Optimization}

Consider a {\it constrained} minimization problem for a general PLQ penalty: 
\begin{equation}
\label{genELQP}
\min_{y\in Y}\quad \rho_{U, M, b, B}(y)
:=
\sup_{u \in U}
\left\{
\left\langle u, b + By\right\rangle - \frac{1}{2}u^TMu
\right\}\;,
\end{equation}
where $Y$ is a polyhedral set, described by 
\begin{equation}
\label{constr}
Y = \{y: A^Ty \leq a\}\;.
\end{equation}
After studying this problem, we will come back to 
consider the estimator  (\ref{probTwo}).

It turns out that a wide class of problems~\eqref{genELQP} are solvable by 
interior point (IP) methods~\cite{KMNY91,NN94,Wright:1997}.
IP methods solve nonsmooth optimization problems 
by working directly with smooth systems of equations characterizing 
the optimality of these problems. 
\cite[Theorem 13]{AravkinBurkePillonetto2013} presents 
a full convergence analysis for IP methods for formulations~\eqref{genELQP}
{without inequality constraints}, so $Y = \R^{N}$in~\eqref{genELQP}. 
While a generalization of the full analysis to cover inequality constraints 
is out of the scope of this paper, we present an important computational 
result showing that  constraints can be included in a straightforward manner, 
and provide the computational complexity of each interior point iteration. 
Moreover, the proof of the result (given in Appendix) shows that 
constraints {\it help the numerical stability} of the interior point iterations. 

\begin{theorem}[Interior Point for PLQ with Constraints]
\label{ipTheorem}
Consider any optimization problem of the form~\eqref{genELQP}, 
with $y \in \mathbb{R}^N$, $b, u\in \mathbb{R}^K$, $C \in \mathbb{R}^{K \times L}$, 
$c\in \mathbb{R}^L$, 
$B \in \mathbb{R}^{K\times N}$, $A \in \mathbb{R}^{N\times P}$, $M \in \mathbb{R}^{K\times K}$,
and $a \in \mathbb{R}^P$.
Suppose that the PLQ satisfies 
\begin{equation}
\label{nullspaceCond}
\mathrm{Null}(M)\cap \mathrm{Null}(C^T) = 0.
\end{equation} 
Suppose also that $M$ contains on the order of $K$ entries, while $C$ contains on the order of $L$ entries.  
Then every interior point iterations can be computed with complexity $O(L + KN^2 + PN^2 + N^3)$. 
\end{theorem}

The assumptions on the structure of $M$ and $C$ 
are satisfied for many common PLQ penalties. For example, 
for $\ell_2$ we have $M = I$ and $C  = 0$, while for $\ell_1$, $M = 0$
and $C$ contains two copies of the identity matrix. 

Turning out attention back to system identification, 
$N = n$ will be the dimension of the impulse response, while $K$ and $L$ 
may depend on $m$; in fact $K \geq m$ always, 
while $L$ depends on the structure of the PLQ penalty. 
To be more specific, we have the following corollary. 

\begin{corollary}
\label{sysIDcor}
Problem~\eqref{probTwo} can be formulated as 
a minimization problem of the form~\eqref{genELQP}. 
If the constraint matrix $A$ has on the order of $n$
entries, while matrices $B$ and $C$ have on the order 
of $m$ entries, each interior point iteration can be solved 
with complexity $O(mn^2 + n^3)$.  
\end{corollary}

Note that the computational complexity of the IP
method scales favorably with the number of measurements $m$ which,  in 
the system identification scenario, is typically much larger than the number of 
unknown impulse response coefficients $n$.

\section{Monte Carlo study}\label{Sec4}

We consider a Monte Carlo study of 300 runs.
At each run, the MATLAB command
\texttt{m=rss(30)} is first used to obtain a SISO continuous-time
system of 30th order. The continuous-time system \texttt{m}
is then sampled at 3 times of its bandwidth, obtaining the
discrete-time system \texttt{md} through  
the commands: \texttt{bw=bandwidth(m); f = bw*3*2*pi;
md=c2d(m,1/f,'zoh')}. If all poles of \texttt{md} are within the
circle with center at the origin and radius 0.95 on the complex
plane, then the feedforward matrix of \texttt{md} is set to 0, 
i.e. \texttt{md.d=0}, and the system is used and saved. \\ 
The system input at each run is white Gaussian noise of unit
variance. The input delay is always equal to 1
and this information is given to every estimator used in the 
Monte Carlo study described below.\\
Data consists of 400 input-output pairs, which 
are collected after getting rid of initial conditions,
and corrupted by a noise generated as a mixture of 
two normals with a fraction of outlier contamination equal to 0.2; i.e.,
$$
e_i \sim 0.8 {\bf{N}}(0,\sigma^2) + 0.2  {\bf{N}}(0,100\sigma^2). 
$$
Here, $\sigma^2$ is randomly
generated in each run as
the variance of the noiseless
output divided by the realization of a random variable 
uniformly distributed on $[1,10]$. With probability 0.2, 
each measurement
may be contaminated by a random error 
whose standard deviation is $10\sigma$.\\
The quality of an estimator is measured 
by computing the fit measure at every run.
To be more specific,
given a generic dynamic system represented by $S(q)$, 
let $\| S(q) \|_2$ denote
the $\ell_2$ norm of its impulse response, numerically computed 
using only the first 100 impulse response coefficients, 
whose mean is denoted by $\bar S(q)$. 
Then, the fit measure for the $j$-th run with estimated model $\hat{G}_j(q)$ is
\begin{equation}\label{eq:fit}
\F_j(G,\hat G_j) = 100 \left(1- \frac{\| G(q) -\hat{G}_j(q) \|_2}{\| G(q) -\bar G(q) \|_2}   \right)
\end{equation}

During the Monte Carlo simulations, the following 5 estimators
are used:

\begin{itemize}
\item  \textit{Oe+oracle}. Classical PEM approach,
    with candidate models given by rational  transfer functions defined by two polynomials  
of the same order. This estimator is implemented
using the \texttt{oe.m} function of the MATLAB System Identification Toolbox 
equipped with the robustification option (\texttt{'LimitError',r})\footnote{As per MATLAB documentation, the value of \texttt{r} 
specifies when to adjust the weight of large errors from quadratic to linear. 
Errors larger than \texttt{r} times the estimated standard deviation have a linear weight in the criteria. 
The standard deviation is estimated robustly as the median of the absolute deviations from the median and divided by 0.7. 
The value \texttt{r=0} disables the robustification and leads to a purely quadratic criterion.} and 
an oracle, which provides a bound
on the best achievable performance of PEM by selecting (at every run)
the model order (between 1 and 20)
and the value of \texttt{r}  ($0,1,2$ or $3$) that maximize (\ref{eq:fit}).

\item  \textit{Oe+CV}. Same as above, except that
\texttt{r=0} (the fit criterion is purely quadratic) and model order
is estimated via cross validation. In particular, data are split into a training and validation data
set of equal size. Then, for every model order ranging from 1 to 20, 
the MATLAB function \texttt{oe.m} (fed with the training set)
is called. The estimate of the order minimizes 
the sum of squared prediction errors 
on the validation set. This is obtained 
by the MATLAB function \texttt{predict.m} 
(imposing null initial conditions) fed with the validation data set.
The final model is computed by \texttt{oe.m}, 
using the estimated value of the order and all the 
available measurements (the 
union of the training and validation sets).

\item  \textit{Oe+CVrob}. Same as above, except that
level of robustification \texttt{r} is also chosen via cross validation
on the grid $\{0,1,2,3\}$.

\item  \textit{SS+$\ell_2$}.  This is the classical stable spline estimator (\ref{eq:MV}),
which uses a quadratic loss and the stable spline regularizer. Hyperparameters 
are determined via marginal likelihood
optimization, as described in subsection \ref{SS+L2}.  
The number of estimated impulse response coefficients,
i.e. the dimension of $x$ in (\ref{MatrixMod}), is $n=100$.
Only the first 100 input-output pairs are used to define the entries of the matrix $H$ in 
(\ref{MatrixMod}), so that the size of the measurement vector $z$ is $m=300$.
\item  \textit{SS+$\ell_1$}. This is the new nonsmooth 
version of the stable spline estimator. It coincides with (\ref{eq:MV})
except that the quadratic loss
is replaced by the $\ell_1$ loss. The hyperparameter $\alpha$
defining the stable spline kernel in (\ref{eq:TC}) and the regularization parameter 
$\gamma$ are estimated via cross validation as follows. 
The matrix $H$ in (\ref{MatrixMod}) is defined as described above.
Then, the remaining 300 input-output pairs are split into a training and validation
data set of equal size. The estimates of the hyperparameters $\alpha,\gamma$ are 
chosen so that the corresponding impulse response estimate (obtained using only the training set)
provides the best prediction on the validation data
(according to a quadratic fit). The candidate hyperparameters are  selected from a 
two-dimensional grid. In particular, $\alpha$ may assume values in
$[0.01,0.05,0.1,0.15,\ldots,0.9,0.95,0.99]$ while $\gamma$ 
varies on a set given by 20 values logarithmically spaced between
$\hat \gamma/100$ and $100 \hat \gamma$, where $\hat \gamma$ is the estimate 
used by \textit{SS+$\ell_2$}.
The final estimate of the impulse
response is computed using the hyperparameter estimates and the union of
 training and validation data sets.  
\end{itemize}

\begin{figure*}
  \begin{center}
   \begin{tabular}{cc}
  \hspace{.1in}
 { \includegraphics[scale=0.57]{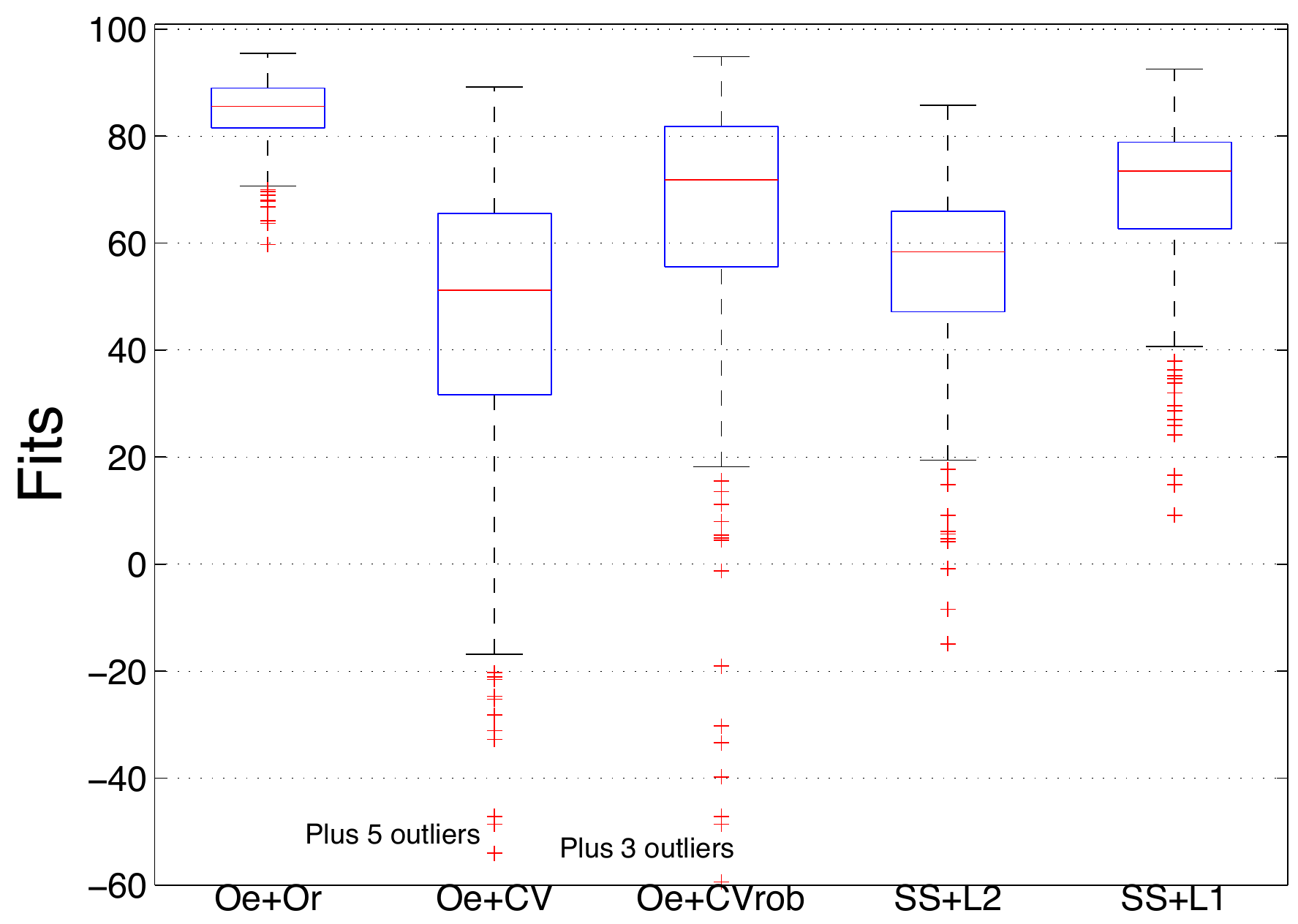}}
   \end{tabular}
 \caption{Boxplot of the 300 percentage fits obtained by PEM equipped with an oracle (\textit{Oe+Or}),
 by PEM with cross validation equipped with the quadratic loss  (\textit{Oe+CV}) and with a robust loss (\textit{Oe+CVrob}),
 by the stable spline estimator equipped with the quadratic loss (\textit{SS+$\ell_2$}) and with the $\ell_1$ loss
 (\textit{SS+$\ell_1$}).} 
    \label{FigOE}
     \end{center}
\end{figure*}

The plots in Fig. \ref{FigOE} are the Matlab
boxplots of the errors (\ref{eq:fit}) obtained by the 5 estimators.
The rectangle shows the $25-75\%$ quantiles of all
the numbers with the horizontal line showing the median.
The ``whiskers" outside the rectangle display the $10-90\%$
quantiles, with the remaining errors (which may be deemed outliers)
plotted using ``+".
The average fits obtained by \textit{Oe+oracle},
\textit{Oe+CV}, \textit{Oe+CVrob},
\textit{SS+$\ell_2$} and \textit{SS+$\ell_1$}
are $84.7,44.4,62.6,55.8$ and $70.1$, respectively.\\
The best results are obtained by \textit{Oe+oracle}. However, keep in mind 
that this estimator relies on an ideal tuning 
of the model order and of the level of robustification which is not implementable in practice.\\  
In comparison with the other estimators, 
the performance of 
\textit{SS+$\ell_2$} and \textit{Oe+CV} is negatively influenced by the presence
of data contamination. The reason is that both of these estimators
use quadratic loss functions. Notice however that  
textit{SS+$\ell_2$} largely outperforms \textit{Oe+CV}.\\ 
Focusing now on numerical schemes equipped with robust losses, we see that \textit{SS+$\ell_1$}
outperforms \textit{Oe+CVrob}. It provides the best results 
among all the estimators implementable in practice: the stable spline kernel 
introduces a suitable regularization with the $\ell_1$ loss
to guard against outliers.

\section{Conclusions}\label{Conclusions}

We have extended the stable spline estimator to a non smooth setting.
Quadratic losses and regularizers can now be replaced by 
general PLQ functions, which allow new applications, such as robust estimators
in the presence of outliers in the data. In addition, we presented 
an extended formulation that can include affine inequality constraints
on the unknown impulse response, which can be used for example to 
incorporate nonnegativity into the estimate. We have shown that 
the corresponding generalized estimates can be computed in an efficient way by IP methods.
Finally, our simulation results showed 
a significant performance improvement of the stable spline kernel with $\ell_1$ loss 
over previous art.

\section{Appendix}

\subsection{Proof of Theorem~\ref{ipTheorem}}

From \cite{RTRW}[Example 11.47], the Lagrangian for
problem (\ref{genELQP}) for feasible $(y, u)$ is given by
\[
L(y, u) = b^Tu -\frac{1}{2}u^TMu + u^TBy\;.
\]
Since $U$ is by assumption a polyhedral set, it can be expressed
by a linear system of inequalities:
\begin{equation}
\label{polyhedralU}
U = \{u: C^Tu \leq c\}\;.
\end{equation}
Using the explicit characterizations of $U$ and $W$, the optimality
conditions for (\ref{genELQP}) are
\begin{equation}
\label{KKTequations}
\begin{aligned}
2By - Mu+ b &= Cq \;, \; q \geq 0\\
-B^Tu &= Aw\;, \; w \geq 0
\end{aligned}
\end{equation}
(see \cite{RTRW} and \cite{RTR} for more details).
The inequality constraint in the definition of $U$
in (\ref{polyhedralU})
can be
reformulated using slack variables $s,r$:
\[
\begin{aligned}
C^Tu + s &= c\\
A^Ty + r &= a\;.
\end{aligned}
\]
Combining all of these equations yields the KKT system for (\ref{genELQP}):
\begin{equation}
\label{fullKKT}
\begin{array}{lll}
0 &=& B^Tu  + Aw\\
0 &=& By - Mu - Cq + b \\
0 &=& C^Tu + s - c\\
0 &=& A^Ty + r - a\\
0 &=& q_is_i \; \forall i\;, \; q, s \geq 0\\
0 &=& w_ir_i \; \forall i\;, \; w, r \geq 0\;.
\end{array}
\end{equation}
The last two sets of equations in (\ref{fullKKT}) are known as the
complementarity conditions. Solving the problem (\ref{genELQP}) is
then equivalent to satisfying (\ref{fullKKT}), and there is a vast
optimization literature on working directly with the KKT system. In
the Kalman filtering/smoothing application, interior point methods
have been used to solve the KKT system (\ref{fullKKT}) in a
numerically stable and efficient manner, see e.g.
\cite{AravkinIFAC}.

An interior point approach applies damped Newton iterations to a relaxed
version of \ref{fullKKT}:
\begin{equation}
\label{rKKT}
F_\mu(s, q, u, r, w, y)
=
\begin{bmatrix}
s + C^Tu - c \\
QS\B{1} - \mu \B{1}\\
By - Mu - Cq + b \\
r + A^Ty - a\\
WR\B{1} - \mu \B{1}\\
B^Tu + Aw
\end{bmatrix}\;.
\end{equation}
The relaxation parameter $\mu$ is driven aggressively to $0$ as the method
proceeds. Every Newton iteration solves
$$
F_\mu^{(1)} \begin{bmatrix} \Delta s^T, & \Delta q^T, & \Delta u^T, & \Delta r^T,&\Delta w^T,& \Delta y^T, \end{bmatrix}^T=
$$
$$
-F_\mu(s, q, u, r, w, y)\;,
$$
where
\begin{equation}
\label{rKKTder}
F_\mu^{(1)}
=
\begin{bmatrix}
I & 0 & C^T & 0 & 0 & 0\\
Q & S & 0 & 0 & 0 & 0\\
0 & -C & -M & 0 & 0 & B\\
0 & 0 & 0 & I & 0 & A^T\\
0 & 0 & 0 & W & R & 0\\
0 & 0 & B^T & 0 & A & 0
\end{bmatrix}\;.
\end{equation}
Using the row operations
\[
\begin{aligned}
r_2 &\gets r_2 - Q r_1\\
r_3 &\gets r_3 + CS^{-1}r_2
\end{aligned}
\]
we arrive at the system
\[
\begin{bmatrix}
I & 0 & C^T & 0 & 0 & 0\\
0 & S & -QC^T & 0 & 0 & 0\\
0 & 0 & -T & 0 & 0 & B\\
0 & 0 & 0 & I & 0 & A^T\\
0 & 0 & 0 & W & R & 0\\
0 & 0 & B^T & 0 & A & 0
\end{bmatrix}\;.
\]
where
$T = M + C\mathrm{diag}(q/s)C^T$.
Note that this matrix is invertible 
if and only if the hypothesis~\eqref{nullspaceCond}
holds. If $T$ is invertible, the row operations
\[
\begin{aligned}
r_6 & \gets r_4 + B^TT^{-1}B\\
r_5 & \gets r_5 - Wr_4\\
r_6 & \gets r_6 - AR^{-1}r_5
\end{aligned}
\]
reduce the system to upper triangular form
\[
\begin{bmatrix}
I & 0 & C^T & 0 & 0 & 0\\
0 & S & -QC^T & 0 & 0 & 0\\
0 & 0 & -T & 0 & 0 & B\\
0 & 0 & 0 & I & 0 & A^T\\
0 & 0 & 0 & 0 & R & -WA^T\\
0 & 0 & 0 & 0 & 0 & B^TT^{-1}B + AR^{-1}WA^T
\end{bmatrix}\;.
\]
Note that $s$, $q$, $r$, and $w$ are componentwise
positive (which holds for every nonzero $\mu$), while $B$ is injective
(see Definition~\ref{generalPLQ}), hence
$B^T T^{-1}B$ is a square matrix of full rank. The term 
$AR^{-1}WA^T$ is also positive semidefinite, and only serves to stabilize the inversion 
of the final term.  
Therefore, we can carry out Newton
iterations on the $\mu$-relaxed system, as claimed.

To show the computational complexity, we give the full interior point iteration,
which is derived by applying the row operations used to obtain the upper 
triangular system to the right hand side $-F_\mu$, 
then solving for $\Delta y$, and back substituting. 
\begin{equation}
\label{IPiter}
\begin{aligned}
r_1 &= -s - C^Tu + c
\\
r_2 &= \mu\B{1} + Q(C^Tu - c)
\\
r_3 &= -(By - Mu - Cq + b) + CS^{-1}r_2
\\
r_4 &= -(r + A^Ty - a)
\\
r_5 &= \mu\B{1} +W(A^Ty - a)
\\
T &= M + CQS^{-1}C^T
\\
r_6 &= -(B^T u + Aw ) +B^T T^{-1}r_3 - AR^{-1}r_5
\\
\Omega &= B^TT^{-1}B + AR^{-1}WA^T
\\
\Delta y &= \Omega^{-1} r_6
\\
\Delta w & = R^{-1}(r_5 + WA^T \Delta y)
\\
\Delta r  & =  r_4 - A^T\Delta y
\\
\Delta u & = T^{-1}(-r_3 + B\Delta y)
\\
\Delta q & = S^{-1}(r_2 + QC^T\Delta u)
\\
\Delta s & = r_1 - C^T \Delta u
\end{aligned}
\end{equation}
Note that the matrix $T$ can be constructed in $O(L + K)$ operations
if $C$ contains on the order of $L$ terms. The matrix $\Omega$
can be constructed in $O(NK^2 + NP^2)$ operations, and inverted 
in $O(N^3)$ operations. These operations dominate the complexity, 
giving the bound $O(L + NK^2 + NP^2 + N^3)$. 

\subsection{Proof of Corollary~\ref{sysIDcor}}

To translate~\eqref{probTwo} to~\eqref{genELQP}, we have to specify the structures 
$A, B, b, C, c$, which capture the impulse response constraints, 
the injective linear model, and the structure of $U$, 
respectively. 

Suppose that $\rho_w(y)$ and $\rho_v(x)$ are given by 
\begin{equation}
\label{VWplqs}
\begin{aligned}
\rho_w(y) &:= \sup_{u \in U_w} \left\langle b_w + B_w y, u\right \rangle - \frac{1}{2}u^T M_w u\\
\rho_v(x)  &:= \sup_{u \in U_v} \left\langle b_v + B_v x, u\right \rangle - \frac{1}{2}u^T M_v u
\end{aligned}
\end{equation}

First define
\[
\begin{aligned}
&\widetilde{\rho}_v(y) := \rho_v(\gamma^{-1}(HLy - z)) \\
& =  
\sup_{u \in U_v} \left\langle b_v - \gamma^{-1}B_v z
+ 
\gamma^{-1}B_v HLy, u\right \rangle - \frac{1}{2}u^T M_v u\;.
\end{aligned}
\]
Adding $\widetilde{\rho}_v$ and $\rho_w$ together, we obtain the general system identification 
objective with the following specification:
\[
\begin{aligned}
M &= \begin{bmatrix} M_w & 0 \\ 0 & M_v \end{bmatrix}, 
\quad 
B = \begin{bmatrix} B_w \\ \gamma^{-1}B_v HL\end{bmatrix}, 
\quad
b = \begin{bmatrix} b_w \\b_v - \gamma^{-1}B_v z\end{bmatrix}
\\
C & = \begin{bmatrix} C_w & 0 \\ 0 & C_v \end{bmatrix},
\quad
c = \begin{bmatrix} c_w \\c_v \end{bmatrix}\;.
\end{aligned}
\]
The matrix $A$ and vector $a$ encodes the constraints,
as given by~\eqref{constr}.

This completes the specification. The complexity result follows
immediately from the assumptions on $A,B,C$ and Theorem~\ref{ipTheorem}.

It is also worthwhile to consider the structure of~\eqref{IPiter}.
First, note that 
\[
\begin{aligned}
T &= M + CQS^{-1}C^T 
\\
&= \begin{bmatrix} M_w & 0 \\ 0 & M_v \end{bmatrix}
+
\begin{bmatrix} C_w & 0 \\ 0 & C_v \end{bmatrix} 
QS^{-1}
\begin{bmatrix} C_w & 0 \\ 0 & C_v \end{bmatrix}^T
\\
&=\begin{bmatrix} M_w + C_w Q_wS^{-1}_WC_w^T& 0 \\ 0 & M_v + C_vQ_vS^{-1}vC_v^T \end{bmatrix}
\\
&=\begin{bmatrix} T_w & 0 \\ 0 & T_v \end{bmatrix}\;,
\end{aligned}
\]
so in fact $T$ is block diagonal. 
This fact gives a more explicit formula for $\Omega$:
\[
\begin{aligned}
\Omega &= B^TT^{-1}B + AR^{-1}WA^T\\
& =  \begin{bmatrix} B_w^T & \gamma^{-1}L^TH^TB_v^T\end{bmatrix} 
\begin{bmatrix} T_w^{-1} & 0 \\ 0 & T_v^{-1} \end{bmatrix}
\begin{bmatrix} B_w \\ \gamma^{-1}B_v HL\end{bmatrix} 
+
AR^{-1}WA^T\\
& = B_w^T T_w^{-1}B_w + \sigma^{-2}L^TH^TB_v^T T_v^{-1}B_v HL + AR^{-1}WA^T\;.
\end{aligned}
\]

\bibliographystyle{IEEEtran}
\bibliography{biblio,biblioSurvey}

\begin{thebibliography}{10}
\providecommand{\url}[1]{#1}
\csname url@rmstyle\endcsname
\providecommand{\newblock}{\relax}
\providecommand{\bibinfo}[2]{#2}
\providecommand\BIBentrySTDinterwordspacing{\spaceskip=0pt\relax}
\providecommand\BIBentryALTinterwordstretchfactor{4}
\providecommand\BIBentryALTinterwordspacing{\spaceskip=\fontdimen2\font plus
\BIBentryALTinterwordstretchfactor\fontdimen3\font minus
  \fontdimen4\font\relax}
\providecommand\BIBforeignlanguage[2]{{%
\expandafter\ifx\csname l@#1\endcsname\relax
\typeout{** WARNING: IEEEtran.bst: No hyphenation pattern has been}%
\typeout{** loaded for the language `#1'. Using the pattern for}%
\typeout{** the default language instead.}%
\else
\language=\csname l@#1\endcsname
\fi
#2}}

\bibitem{Ljung}
L.~Ljung, \emph{System Identification, Theory for the User}.\hskip 1em plus
  0.5em minus 0.4em\relax Prentice Hall, 1999.

\bibitem{Soderstrom}
T.~S{\"o}derstr{\"o}m and P.~Stoica, \emph{System Identification}.\hskip 1em
  plus 0.5em minus 0.4em\relax Prentice-Hall, 1989.

\bibitem{Akaike1974}
H.~Akaike, ``A new look at the statistical model identification,'' \emph{IEEE
  Transactions on Automatic Control}, vol.~19, pp. 716--723, 1974.

\bibitem{Hastie01}
T.~J. Hastie, R.~J. Tibshirani, and J.~Friedman, \emph{The Elements of
  Statistical Learning. Data Mining, Inference and Prediction}.\hskip 1em plus
  0.5em minus 0.4em\relax Canada: Springer, 2001.

\bibitem{SS2010}
G.~Pillonetto and G.~{De Nicolao}, ``A new kernel-based approach for linear
  system identification,'' \emph{Automatica}, vol.~46, no.~1, pp. 81--93, 2010.

\bibitem{PitfallsCV12}
------, ``Pitfalls of the parametric approaches exploiting cross-validation or
  model order selection,'' in \emph{Proceedings of the 16th IFAC Symposium on
  System Identification (SysId 2012)}, 2012.

\bibitem{SS2011}
G.~Pillonetto, A.~Chiuso, and G.~D. Nicolao, ``Prediction error identification
  of linear systems: a nonparametric {G}aussian regression approach,''
  \emph{Automatica}, vol.~47, no.~2, pp. 291--305, 2011.

\bibitem{Rasmussen}
C.~Rasmussen and C.~Williams, \emph{{G}aussian Processes for Machine
  Learning}.\hskip 1em plus 0.5em minus 0.4em\relax The MIT Press, 2006.

\bibitem{ChenOL12}
T.~Chen, H.~Ohlsson, and L.~Ljung, ``On the estimation of transfer functions,
  regularizations and {G}aussian processes - revisited,'' \emph{Automatica},
  vol.~48, no.~8, pp. 1525--1535, 2012.

\bibitem{PillACC2010}
G.~Pillonetto, A.~Chiuso, and G.~{De Nicolao}, ``Regularized estimation of sums
  of exponentials in spaces generated by stable spline kernels,'' in
  \emph{Proceedings of the IEEE American Cont. Conf., Baltimora, USA}, 2010.

\bibitem{Maritz:1989}
J.~S. Maritz and T.~Lwin, \emph{Empirical Bayes Method}.\hskip 1em plus 0.5em
  minus 0.4em\relax Chapman and Hall, 1989.

\bibitem{MacKay}
D.~MacKay, ``Bayesian interpolation,'' \emph{Neural Computation}, vol.~4, pp.
  415--447, 1992.

\bibitem{BergerBook}
J.~Berger, \emph{Statistical Decision Theory and Bayesian Analysis}, 2nd~ed.,
  ser. Springer Series in Statistics.\hskip 1em plus 0.5em minus 0.4em\relax
  Springer, 1985.

\bibitem{AravkinIFAC12}
A.~Aravkin, J.~Burke, and G.~Pillonetto, ``A statistical and computational
  theory for robust and sparse kalman smoothing,'' in \emph{Proceedings of the
  16th IFAC Symposium on System Identification (SysId 2012)}, 2012.

\bibitem{Hub}
P.~Huber, \emph{Robust Statistics}.\hskip 1em plus 0.5em minus 0.4em\relax
  Wiley, 1981.

\bibitem{Gao2008}
J.~Gao, ``Robust l1 principal component analysis and its {B}ayesian variational
  inference,'' \emph{Neural Computation}, vol.~20, no.~2, pp. 555--572,
  February 2008.

\bibitem{Aravkin2011tac}
A.~Aravkin, B.~Bell, J.~Burke, and G.~Pillonetto, ``An $\ell_1$-laplace robust
  kalman smoother,'' \emph{Automatic Control, IEEE Transactions on}, vol.~56,
  no.~12, pp. 2898--2911, dec. 2011.

\bibitem{Farahmand2011}
S.~Farahmand, G.~Giannakis, and D.~Angelosante, ``Doubly robust smoothing of
  dynamical processes via outlier sparsity constraints,'' \emph{IEEE
  Transactions on Signal Processing}, vol.~59, pp. 4529--4543, 2011.

\bibitem{Hastie90}
T.~J. Hastie and R.~J. Tibshirani, ``Generalized additive models,'' in
  \emph{Monographs on Statistics and Applied Probability}.\hskip 1em plus 0.5em
  minus 0.4em\relax London, UK: Chapman and Hall, 1990, vol.~43.

\bibitem{LARS2004}
B.~Efron, T.~Hastie, L.~Johnstone, and R.~Tibshirani, ``Least angle
  regression,'' \emph{Annals of Statistics}, vol.~32, pp. 407--499, 2004.

\bibitem{Donoho2006}
D.~Donoho, ``Compressed sensing,'' \emph{IEEE Trans. on Information Theory},
  vol.~52, no.~4, pp. 1289--1306, 2006.

\bibitem{Lasso1996}
R.~Tibshirani, ``Regression shrinkage and selection via the {LASSO},''
  \emph{Journal of the Royal Statistical Society, Series B.}, vol.~58, pp.
  267--288, 1996.

\bibitem{EN_2005}
H.~Zou and T.~Hastie, ``Regularization and variable selection via the elastic
  net,'' \emph{Journal of the Royal Statistical Society, Series B}, vol.~67,
  pp. 301--320, 2005.

\bibitem{Vapnik98}
V.~Vapnik, \emph{Statistical Learning Theory}.\hskip 1em plus 0.5em minus
  0.4em\relax New York, NY, USA: Wiley, 1998.

\bibitem{Pontil98}
M.~Pontil and A.~Verri, ``Properties of support vector machines,'' \emph{Neural
  Computation}, vol.~10, pp. 955--974, 1998.

\bibitem{Evgeniou99}
T.~Evgeniou, M.~Pontil, and T.~Poggio, ``Regularization networks and support
  vector machines,'' \emph{Advances in Computational Mathematics}, vol.~13, pp.
  1--150, 2000.

\bibitem{Scholkopf00}
B.~Sch\"{o}lkopf, A.~J. Smola, R.~C. Williamson, and P.~L. Bartlett, ``New
  support vector algorithms,'' \emph{Neural Computation}, vol.~12, pp.
  1207--1245, 2000.

\bibitem{AravkinCDC12}
A.~Aravkin, J.~Burke, and G.~Pillonetto, ``Nonsmooth regression and state
  estimation using piecewise quadratic log-concave densities,'' in
  \emph{Proceedings of the 51st IEEE Conference on Decision and Control (CDC
  2012)}, 2012.

\bibitem{AravkinBurkePillonetto2013}
A.~Y. Aravkin, J.~V. Burke, and G.~Pillonetto, ``Sparse/robust estimation and
  kalman smoothing with nonsmooth log-concave densities: Modeling,computation,
  and theory, 2013.''

\bibitem{CDC2011P1}
G.~Pillonetto and G.~{De Nicolao}, ``Kernel selection in linear system
  identification -- part {I}: A {G}aussian process perspective,'' in
  \emph{Proceedings of CDC-ECC}, 2011.

\bibitem{CDC2011P2}
T.~Chen, H.~Ohlsson, G.~Goodwin, and L.~Ljung, ``Kernel selection in linear
  system identification -- part {II}: A classical perspective,'' in
  \emph{Proceedings of CDC-ECC}, 2011.

\bibitem{Goodwin1992}
G.~Goodwin, M.~Gevers, and B.~Ninness, ``Quantifying the error in estimated
  transfer functions with application to model order selection,'' \emph{IEEE
  Transactions on Automatic Control}, vol.~37, no.~7, pp. 913--928, 1992.

\bibitem{RTRW}
R.~Rockafellar and R.~Wets, \emph{Variational Analysis}.\hskip 1em plus 0.5em
  minus 0.4em\relax Springer, 1998, vol. 317.

\bibitem{Chu01aunified}
W.~Chu, S.~S. Keerthi, and C.~J. Ong, ``A unified loss function in bayesian
  framework for support vector regression,'' in \emph{In Proceeding of the 18th
  International Conference on Machine Learning}, 2001, pp. 51--58.

\bibitem{KMNY91}
M.~Kojima, N.~Megiddo, T.~Noma, and A.~Yoshise, \emph{A Unified Approach to
  Interior Point Algorithms for Linear Complementarity Problems}, ser. Lecture
  Notes in Computer Science.\hskip 1em plus 0.5em minus 0.4em\relax Berlin,
  Germany: Springer Verlag, 1991, vol. 538.

\bibitem{NN94}
A.~Nemirovskii and Y.~Nesterov, \emph{Interior-Point Polynomial Algorithms in
  Convex Programming}, ser. Studies in Applied Mathematics.\hskip 1em plus
  0.5em minus 0.4em\relax Philadelphia, PA, USA: SIAM, 1994, vol.~13.

\bibitem{Wright:1997}
S.~Wright, \emph{Primal-dual interior-point methods}.\hskip 1em plus 0.5em
  minus 0.4em\relax Englewood Cliffs, N.J., USA: Siam, 1997.

\bibitem{RTR}
R.~Rockafellar, \emph{Convex Analysis}, ser. Priceton Landmarks in
  Mathematics.\hskip 1em plus 0.5em minus 0.4em\relax Princeton University
  Press, 1970.

\bibitem{AravkinIFAC}
A.~Aravkin, B.~Bell, J.~Burke, and G.~Pillonetto, ``Learning using state space
  kernel machines,'' in \emph{Proc. IFAC World Congress 2011}, Milan, Italy,
  2011.

\end{thebibliography}

\end{document}